\pdfoutput=1
\documentclass[11pt]{article}

\usepackage[final]{naacl2021}

\usepackage{times}
\usepackage{latexsym}

\usepackage[T1]{fontenc}
\usepackage[utf8]{inputenc}

\usepackage{microtype}

\usepackage{amsmath}
\usepackage{booktabs}
\usepackage{makecell}
\usepackage{multirow}
\usepackage{pifont}
\usepackage{multicol}
\usepackage{amssymb}
\usepackage{latexsym}
\usepackage{graphicx}
\usepackage{ulem}
\usepackage{tikz}
\usepackage{pgfplots}
\usepackage{tikz-dependency}
\usepackage{subcaption}

\usepackage{microtype}
\usepackage[switch]{lineno}

\title{From Fully Trained to Fully Random Embeddings: Improving Neural Machine Translation with Compact Word Embedding Tables}

\author{Krtin Kumar\thanks{*Work done while Krtin Kumar, Peyman Passban and Yiu Sing Liu were at Huawei. } \\
  Thomson Reuters \\
  \small \texttt{krtinkumar@gmail.com} \\\And
  Peyman Passban$^*$\\
  Amazon \\
  \small \texttt{passban.peyman@gmail.com} \\\And
  Mehdi Rezagholizadeh \\
  Huawei Noah's Ark Lab \\
\AND
  Yiu Sing Lau$^*$  \\
  McGill University, Statistics \\
  \small \texttt{yiusinglau17@gmail.com} \\\And
  Qun Liu \\
  Huawei Noah's Ark Lab \\
  \small \texttt{qun.liu@huawei.com}
 }

\begin{document}
\maketitle
\begin{abstract}
Embedding matrices are key components in neural natural language processing (NLP) models that are responsible to provide numerical representations of input tokens.\footnote{In this paper words and subwords are referred to as \textit{tokens} and the term \textit{embedding} only refers to embeddings of inputs.} In this paper, we analyze the impact and utility of such matrices in the context of neural machine translation (NMT). We show that detracting syntactic and semantic information from word embeddings and running NMT systems with random embeddings is not as damaging as it initially sounds. We also show how incorporating only a limited amount of task-specific knowledge from fully-trained embeddings can boost the performance NMT systems. Our findings demonstrate that in exchange for negligible deterioration in performance, any NMT model can be run with partially random embeddings. Working with such structures means a minimal memory requirement as there is no longer need to store large embedding tables, which is a significant gain in industrial and on-device settings. We evaluated our embeddings in translating {English} into {German} and {French} and achieved a $5.3$x compression rate. Despite having a considerably smaller architecture, our models in some cases are even able to outperform state-of-the-art baselines. 
\end{abstract}

\section{Introduction}
One of the main challenges in NLP is to properly encode discrete tokens into continuous vector representations. The most common practice in this regard is to use an embedding matrix which provides a one-to-one mapping from tokens to $n$-dimensional, real-valued vectors \citep{mikolov2013distributed,li2018word}. Typically, values of these vectors are optimized via back-propagation with respect to a particular objective function. Learning embedding matrices with robust performance across different domains and data distributions is a complex task, and can directly impact quality \cite{tian2014probabilistic,shi2015learning,sun2016sparse}. 

% Added the below para
Embedding matrices are a key component in a seq2seq NMT model, for instance, in a transformer-base NMT model \cite{vaswani2017attention} with a vocabulary size of $50k$, $36\%$ of the total model parameters are utilized by embedding matrices. Thus, a significant amount of parameters are utilized only for representing tokens in a model. Existing work on embedding compression for NMT systems \cite{khrulkov2019tensorized,chen2018groupreduce,shu2017compressing}, have shown that these matrices can by compressed significantly with minor drop in performance. Our focus in this work is to study the importance of embedding matrices in the context of NMT. We experiment with Transformers \cite{vaswani2017attention} and LSTM based seq2seq architectures, which are the two most popular seq2seq models in NMT. In Section \ref{sec:rwe}, we compare the performance of a fully-trained embedding matrix, with a completely random word embedding (RWE), and find that using a random embedding matrix leads to drop in about $1$ to $4$ BLEU \cite{papineni2002bleu} points on different NMT benchmark datasets. Neural networks have shown impressive performance with random weights for image classification tasks \cite{ramanujan2020s}, our experiments show similar results for embedding matrices of NMT models.

\begin{figure*}[t]
  \centering
  \includegraphics[width=0.8\linewidth]{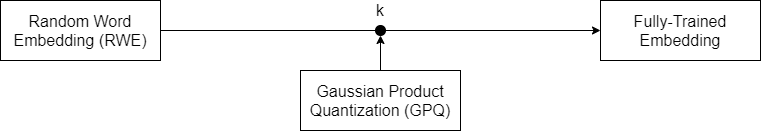}
  \caption{$k$ is a hyper-parameter that controls the number of Gaussian distributions required to approximate the embedding matrix.}
  \label{fig:RWEvsFully}
\end{figure*}

RWE uses no trainable parameters, thus it might be possible to recover the drop in BLEU score by increasing the number of parameters using additional layers. We increase the number of layers in RWE model such that the number of parameters used by the entire model is comparable to fully-trained embedding model. We find that even in comparable parameter setting RWE model performance was inferior to fully-trained model by $~1$ BLEU point for high and medium resource datasets. Our results suggest that even though the embedding parameters can be compressed to a large extent with only a minor drop in accuracy, embedding matrices are essential components for token representation, and cannot be replaced by deeper layers of transformer based models.

RWE assumed a fully random embedding sampled from a single Gaussian distribution; however, to control the amount of random information, and to better understand the importance of embedding matrices, we introduce \textit{Gaussian product quantization} (GPQ). GPQ assumes that $k$ Gaussian distributions are required to approximate the embedding matrix for NMT models. The means and variances of the $k$ distributions are learned from a fully-trained embedding matrix trained on the same dataset as GPQ model. $k$ is a hyper-parameter which controls the amount of information distilled from a pre-trained embedding to partially random embedding in GPQ model. GPQ has the ability to move from a fully random embedding to a fully-trained embedding by increasing the number of distributions $k$, as shown in Figure \ref{fig:RWEvsFully}. Our results show that only $50$ Gaussian distributions are sufficient to approximate the embedding matrix, without drop in performance. GPQ compresses the embedding matrix $5$ times without any significant drop in performance, and uses only $100$ floating point values for storing the means and variances. GPQ demonstrates effective regularization of the embedding matrix, by out-performing the transformer baseline model with fully-trained embedding by $1.2$ BLEU points for $En \rightarrow Fr$ dataset and $0.5$ BLEU for $Pt \rightarrow En$ dataset. A similar increase in performance was also observed for LSTM based models, thus further showing the effectiveness of GPQ for embedding regularization.

% para to explain it is linked to PQ, how other research in this field have not outperformed the baseline.
Product Quantization (PQ) \citep{jegou2010product} was proposed for fast search by approximating nearest neighbour search. PQ has been adapted for compressing embedding matrices for different NLP problems \cite{shu2017compressing,kim2020adaptive,tissier2019near,li2018slim}. Our method is an extension of PQ, first, we incorporate variance information in GPQ, then we define \textit{unified partitioning} to learn a more robust shared space for approximating the embedding matrix. Our extensions consistently outperform the original PQ-based models with better compression rates and higher BLEU scores. These improvement are observed for Transformer-based models as well as conventional, recurrent neural networks.

% \sout{Based on the aforementioned experiments, we could work with random embeddings and perform relatively on par with the baseline model. To close the performance gap we placed extra projection layers after the embedding table. In another effort, we tried to eliminate new parameters introduced with these additional layers and yet be comparable to the baseline, so we utilized product quantization (PQ).}

% \sout{We first trained a translation model and applied PQ \citep{jegou2010product} to its embeddings. PQ decomposes the embedding table into two considerably smaller tables by which we can reduce memory consumption. Original embeddings are still reversible using these two new matrices. This simple modification reduces the embedding matrix size by $5.3$ times with almost no information loss in our translation engines. For the \textit{Pt}$\rightarrow$\textit{En} and  \textit{En}$\rightarrow$\textit{Fr} directions it even improves the BLEU score by $1.2$ points.} 

% \sout{In addition to applying PQ on the embedding matrix, we propose two other extensions, referred to as \textit{Gaussian product quantization} (GPQ) and \textit{unified partitioning}. Our extensions consistently outperform the original PQ-based models with better compression rates and higher BLEU scores. These improvement are observed for Transformer-based models as well as conventional, recurrent neural networks.}

\section{Product Quantization (PQ)}
\label{sec:background}
\begin{figure}[h]
  \centering
  \includegraphics[width=0.8\linewidth]{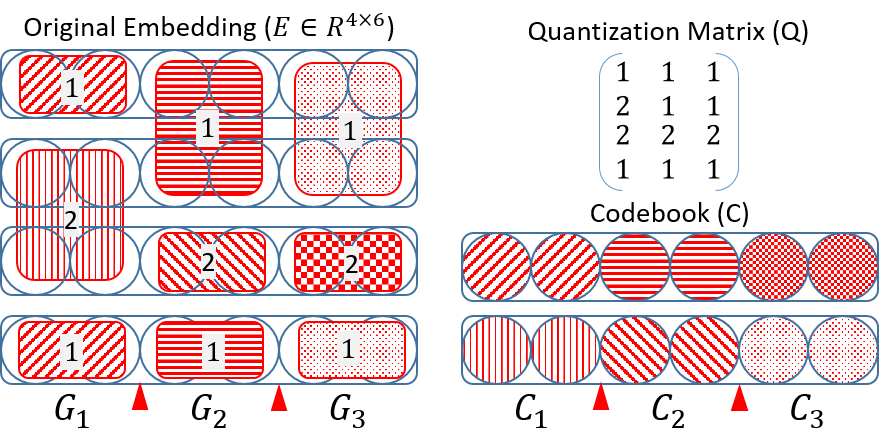}
  \caption{The matrix on the left hand side shows the original embedding matrix $E$, after dividing columns into $3$ groups ($G_i; i\in\{1,2,3\}$) and applying the k-means algorithm with the cluster number $2$ ($c=2$) to each group. The digit inside each block is the cluster number. The figure on the right hand side shows the quantization matrix Q, and codebook C.}
  \label{fig:pqsimple}
\end{figure}

PQ is the core of our techniques, so we briefly discuss its details in this section. For more details see \citet{jegou2010product}. As previously mentioned, NLP models encode tokens from a discrete to a continuous domain via an embedding matrix \textit{E}, as simply shown in Equation \ref{eqn:wordembedding}:
\begin{equation}
    E \in \mathbb{R}^{|V| \times n}
  \label{eqn:wordembedding}
\end{equation}
where $V$ is a vocabulary set of unique tokens and $n$ is the dimension of each embedding. We use the notation $E_{w} \in \mathbb{R}^n$ to refer to the embedding of the \textit{w}-th word in the vocabulary set. In PQ, first $E$ is partitioned into $g$ groups across columns with $G_{i} \in \mathbb{R}^{|V| \times \frac{n}{g}}$ representing the \textit{i}-th group. Then each group $G_{i}$ is clustered using K-means into $c$ clusters. Cluster indices and cluster centroids are stored in the vector $Q_{i} \in \mathbb{N}^{|V|}$ and matrix $C_{i} \in \mathbb{R}^{c \times \frac{n}{g}}$, respectively. Figure \ref{fig:pqsimple} illustrates the PQ-based decomposition process.

We can obtain the {quantization matrix} $Q$ (also known as the \textit{index matrix} in the literature) and \textit{codebook} $C$ by concatenating $Q_i$ and $C_i$ for all $g$ groups along the columns, as shown in Equation \ref{eqn:pqconcat}.     

\begin{equation}
    \begin{split}
        & Q = \text{Concat}_{\text{column}}(Q_1, Q_2, ..., Q_g) \\
        & C = \text{Concat}_{\text{column}}(C_1, C_2,...,C_g)
    \end{split}
    \label{eqn:pqconcat}
\end{equation}
%and $v_f \in \mathbb{R}^{|V|}$ to represent the \textit{f}-th column. 
%\textcolor{red}{Notation inconsistency: Are we showing vectors with small letters and matricies with big letters? Then we need to change $E_w$ to $e_w$.}

The decomposition process applied by PQ is reversable, namely any embedding table decomposed by PQ can be reconstructed using the matrices $Q_{i}$ and $C_{i}$. The size of an embedding table compressed via PQ can be calculated as shown in Equation \ref{eqn:PQparamsize}: 
\begin{equation} 
    \text{Size}_{\text{PQ}} = \text{log}_2(c)\times|V|\times g + c n f_p \text{ (in bits)}
    \label{eqn:PQparamsize}
\end{equation}
where $f_p$ is the number of bits defined to store parameters with floating points. In our experiments $f_p$ is $32$ for all settings. We use Equation \ref{eqn:PQparamsize} to measure the compression rate of different models. 

\section{Methodology} 
In this section we first explain RWE which stands for Random Word Embeddings. In RWE, all embeddings are initialized with completely random values where there is no syntactic and  semantic information is available.

Next, we move from completely random embeddings to weakly supervised embeddings by incorporating some knowledge from a pre-trained embedding matrix. We propose our Gaussian Porduct Quantization (GPQ) technique in this regard, which applies PQ to a pre-trained embedding matrix and models each cluster with a Gaussian distribution. Our GPQ empowers the PQ with taking intra-cluster variance information into account. This approach is particularly useful when PQ clusters have a high variance, which might be the case for large embedding matrices.

\subsection{Random Word Embeddings (RWE)}
\label{sec:rwe}
A simple approach to form random embeddings is to sample their values from a standard Gaussian distribution, as shown in Equation \ref{eqn:univariate}: 
% We propose to encode words to continuous space vectors, and assume that encoded vectors do not contain any semantic or syntactic information. Our assumption allows us to establish a baseline performance of our model, allowing us to analyze the utility of embedding matrix. Under the assumption of no semantic or syntactic information, a reasonable way of distributing vectors in $n$-dimensional hyper-sphere would be to ensure that each embedding-vector is equidistant from rest of the embedding vectors in the vocabulary. 

% This can be achieved by minimizing Equation \ref{eqn:equidistant}, 
% \begin{equation}
    % \min \theta(E) = \max_{||E_k||=1} \left \{ \min_{E_i \in E} (arccos(E_k \cdot E_i)) \right \}
    % \label{eqn:equidistant}
% \end{equation}
% Solving Equation \ref{eqn:equidistant} is an NP-hard problem, though different approximations to the problem exists \cite{lovisolo2001uniform}. The simplest and most efficient approximation to Equation \ref{eqn:equidistant} is to sample from a standard Gaussian distribution and normalize the embedding vectors as shown in Equation \ref{eqn:univariate},
\begin{align}
    S &= \mathcal{N}(E_{i,j}| \mu, \sigma^2) \; \forall (i,j) \nonumber \\
    E^{'}_i &= \frac{S_{i}}{||S_{i}||} \; \forall i, 
    \label{eqn:univariate}
\end{align}
where $\mathcal{N}$ is a normal distribution with $\mu=0$ and $\sigma=1$, $S$ is sampled matrix of the same dimensions as $E$, $E'$ is the reconstructed matrix, $E_i$ and $S_i$ represent $i^{th}$ row vector of matrices $E'$ and $S$ respectively.
We normalize each word embedding vector to unify the magnitude of all word embeddings to ensure that we do not add any bias to the embedding matrix. 

Since this approach relies on completely random values, to increase the expressiveness of embeddings and handle any potential dimension mismatch between the embedding table and the first layer of the neural model we place an \textit{optional} linear transformation matrix $W$ (where $W \in \mathbb{R}^{n \times m}$). The weight matrix $W$ uses $n \times m$ trainable parameters, where $n$ is the size of embedding vector and $m$ is the dimension in which the model accepts its inputs. The increase in the number of parameters is negligible as it only relies on the embedding and input dimensions ($n$ and $m$), which are both considerably smaller than the number of words in a vocabulary ($n,m \ll |V|$).

% \subsection{K-Gaussian Estimate}
\subsection{Gaussian Product Quantization (GPQ)}
\label{sec:gpq}
RWE considers completely random embeddings which can be a very strict condition for an NLP model. Therefore, we propose our Gaussian Product Quantization (GPQ) to boost random embeddings with prior information from a pre-trained embedding matrix. 
%our altered product quantization approach which we call it Gaussian Product Quantization (GPQ).
We assume that there is an existing embedding matrix $E$ that is obtained from a model with the same architecture as ours, trained for the same task using the same dataset. 

The pipeline defined for GPQ is as follows: First, we apply PQ to the pre-trained embedding matrix $E$ to derive the quantization matrix $Q_{|V|\times g}$ and the codebook matrix $C_{c\times n}$. The codebook matrix stores centers of $c$ clusters for all $G_i$ groups and the quantization matrix stores the mapping index of each sub-vector in the embedding matrix to the corresponding center of the cluster in the codebook.  The two matrices $Q$ and $C$ can be used to reconstruct the original matrix $E$ from the PQ process.   

\begin{figure*}[t]
  \centering
  
  \begin{subfigure}[t]{.24\linewidth}
    \includegraphics[width=1.0\linewidth]{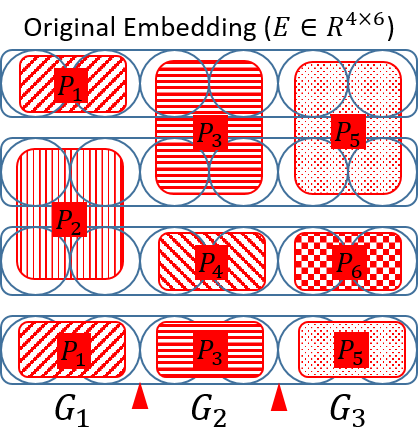}
    \caption{\label{fig:orig}}
  \end{subfigure}
  \begin{subfigure}[t]{.45\linewidth}
    \includegraphics[width=1.0\linewidth]{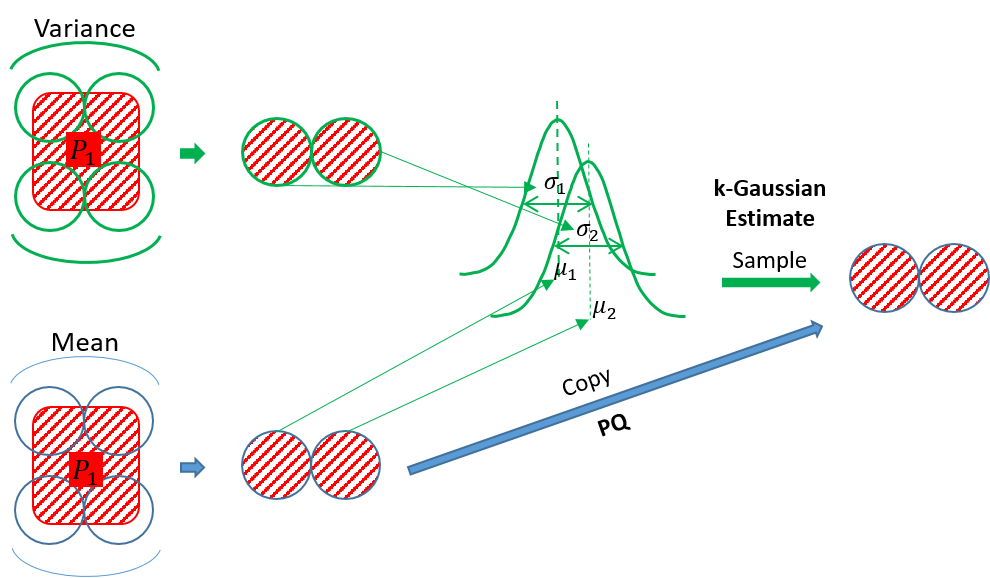}
    \caption{\label{fig:comp}}
  \end{subfigure}
  \begin{subfigure}[t]{.24\linewidth}
    \includegraphics[width=1.0\linewidth]{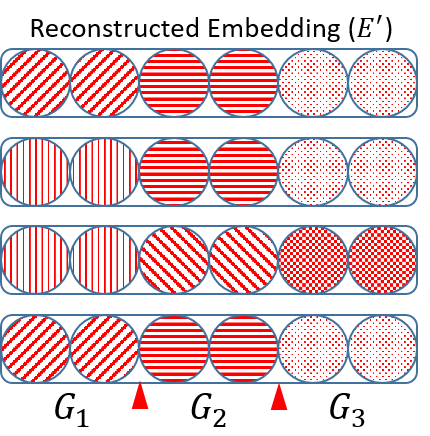}
    \caption{\label{fig:rec}}
  \end{subfigure}
  \caption{\label{fig:main}The figure on the left hand side shows the original embedding matrix $E$ after applying the k-means algorithm, with 2 clusters ($c=2$) and 3 groups ($G_i$) ($g=3$), which results in 6 partitions ($P_i$). The figure in the middle shows the difference between \textit{PQ} and GPQ. The figure on the right hand side is the reconstructed matrix $E^{'}$.}
\end{figure*}

The PQ technique only stores the center of clusters in the codebook and does not take variance of them into account. However, our GPQ technique models each cluster with a single Gaussian distribution by setting its  mean and variance parameters to the cluster center and intra-cluster variance. We define $\mathcal{C}_i^j , (1 \leq j \leq c)$ to be the cluster corresponding to the $j-{\text{th}}$ row of $C_i$  in the codebook (that is $C_i^j$) with the mean and variance of $\mu_i^j \in \mathbb{R}^{\frac{c}{n}}$ and $(\sigma_i^j)^2$. Then, we define entries of the codebook of our GPQ approach as following:
\begin{equation}
    \begin{split}
        & \hat{C}_i^j \sim \mathcal{N}_i^j (\mu_i^j,(\sigma_i^j)^2).
    \end{split}
    \label{GPQ:Gaussaian}
\end{equation}
%Consequently, we would have a non-deterministic version of PQ where the $C_i$ elements of the codebook matrix can be sampled from the defined Gaussian distributions in Equation~\ref{GPQ:Gaussaian}.
Consequently, we model each cluster of PQ with a single Gaussian distribution with two parameters.  
Then the embedding matrix $\hat{E}$ can be reconstructed using a lookup function that maps values $Q_i$ from $\hat{C}_i$. We illustrate our GPQ method and compare it with PQ in Figure \ref{fig:main}. 
In GPQ, we only need to store the index matrix, codebook and their corresponding variances in order to be able to reconstruct the embedding matrix.

The PQ/GPQ method relies on partitioning the embedding matrix into same size $G_i$ groups across the columns and then clustering them. We refer to the partitioning function used in PQ as \textit{Structured Partitioning}, and propose a new partitioning scheme, which we refer to as \textit{Unified Partitioning}. The details of each scheme is described in the following.

\subsubsection{Structured Partitioning}
\label{sec:structured}

The original PQ method is based on structured partitioning to partition the input matrix $E_{|V|\times n}$ into $g$ groups of $G_i$ of size $|V|\times \frac{n}{g}$ along the columns uniformly such that 
$E=\text{Concat}_{\text{column}}(G_1, G_2, ..., G_g)$. Each $G_i$ group is  clustered into $c$ clusters along the rows using any clustering algorithm. If we choose K-means clustering algorithm for this purpose, then we can obtain the center of clusters $C_i  \subset C \text{ (where } C_i \in \mathbb{R}^{c\times \frac{n}{g}})$, their variances $\sigma_i^2 \in \mathbb{R}^{{c}}$, and the quantization vector $Q_i \in \mathbb{N^{|V|}}$ corresponding to $G_i$ as following: 

\begin{equation}
    \begin{split}
        & [C_i,\sigma_i^2, Q_i] = \text{K-means}(G_i,c).
    \end{split}
\end{equation}

The total number of clusters in this case is $k=c*g$. In our technique, we store the variance of clusters in addition to their mean which utilizes more number of floating point parameters compared to Equation \ref{eqn:PQparamsize}, as shown in Equation \ref{eqn:SPQparamsize}.
\begin{equation}
    \text{Size}_{\text{GPQ}}^{\text{Structured}} = \text{log}_2(c) \times|V| \times g + 2 c n f_p \text{ bits} 
    \label{eqn:SPQparamsize}
\end{equation}

% The \textit{structured} partitioning function $F_s$ is used in PQ to partition a 2-dimensional matrix, we formally define it in Equation \ref{eqn:structured}. $F_s$ divides the embedding dimension $n$ into $g$ groups. Each of these groups are then clustered into $c$ clusters along the vocabulary dimension of size $|V|$. We can choose any hard clustering algorithm for $F_c$. We chose K-means clustering as our clustering algorithm. 
% \begin{align}
%      F_s(c,g) &= \{F_c(G_1) \cup \cdots \cup F_c(G_g)\} \nonumber \\
%      J(j, i) &= c*(i-1)+j  \nonumber \\
%      F_c(G_i) &= \{(Q_{J(j,i)}, P_{J(j,i)}) \; \forall j=1\dots c\},
%     \label{eqn:structured}
% \end{align}
% where $c$ and $g$ are hyper-parameters, $G_i$ represent the $i^{th}$ group. $Q$ and $P$ represents sets of quantization matrices and partitions respectively. The function $J(j,i)$ is used to access the quantization matrix and partition corresponding to the $i^{th}$ group and $j^{th}$ cluster. The total number of partitions in this case is $k=c*g$. In addition to the mean we also store the variance of each partition, thus our method uses two times more floating point parameters compared to Equation \ref{eqn:PQparamsize}, as shown in Equation \ref{eqn:SPQparamsize}.
% \begin{equation}
%     size = log_2(c)|V|g + 2 c n f_p \text{ } bytes
%     \label{eqn:SPQparamsize}
% \end{equation}

\subsubsection{Unified Partitioning}
\label{sec:unified}

\begin{figure}
  \centering
  \includegraphics[width=1.0\linewidth]{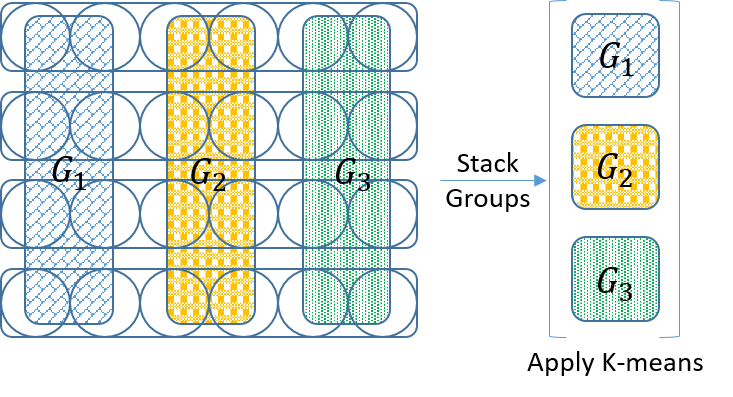}
  \caption{\label{fig:unifiedPQ} \textit{Unified} partitioning function with 3 Groups.}
\end{figure}

\textit{Structured} partitioning has limited ability to exploit redundancies across groups. Motivated by this shortcoming we propose our \textit{unified} partitioning approach in which we concatenate the $g$ groups along rows to form a matrix $\mathcal{G}\in \mathbb{R}^{g|V|\times \frac{n}{g}}$ and then apply K-Means to the matrix $\mathcal{G}$ (instead of applying K-means to each group $G_i$ separately).  
\begin{align}
     \mathcal{G} &= \text{Concat}_{row}(G_{1} , G_{2} , ..., G_{g}) \in \mathbb{R}^{(g|V|) \times (n/g)} \\
     & [C,\sigma^2, Q] = \text{K-means}(\mathcal{G} ,c).
    \label{eqn:unified}
\end{align}
% In this case, the number of partitions is equal to the number of clusters ($k=c$). 
For better understanding we illustrate our method in Figure \ref{fig:unifiedPQ}. Equation \ref{eqn:USPQparamsize} defines the size of parameters for \textit{unified} partitioning function.
\begin{equation}
    \text{Size}_{\text{GPQ}}^{\text{Unified}} = \text{log}_2(c)\times |V| \times g + 2 c \frac{n}{g} f_p \text{ bits}
    \label{eqn:USPQparamsize}
\end{equation}

% \textit{Structured} partitioning has limited ability to exploit redundancies across groups. Motivated by this shortcoming we propose our \textit{unified} partitioning approach $F_u$, as shown in Equation \ref{eqn:unified}, 
% \begin{align}
%      U &= G_{1} || G_{2} || \cdots G_{g} \in \mathbb{R}^{|V|g \times n/g} \\
%      F_u &= F_c(U) = \{(Q_{i}, P_{i}) \forall i=1\dots c\},
%     \label{eqn:unified}
% \end{align}
% where $U$ is the unified matrix to which the clustering algorithm $F_c$ is applied, $G_i$ represent the $i^{th}$ group, and $|\cdot|$ is the concatenation operator. We again choose K-means as our clustering algorithm. The number of partitions is equal to the number of clusters in this case ($k=c$). For better understanding we illustrate our method in Figure \ref{fig:unifiedPQ}. Equation \ref{eqn:USPQparamsize} defines the size of parameters for $unified$ partitioning function.
% \begin{equation}
%     size = log_2(c)\times |V|g + 2 c \frac{n}{g} f_p \text{ } bytes
%     \label{eqn:USPQparamsize}
% \end{equation}

\begin{table}[t]
\centering
\begin{tabular}{|l|c|c|}
\toprule
Model & Float & Integer \\ \hline
PQ & 25.6k & 16.3M \\
PQ (Unified)  & 50 & 16.3M \\
GPQ (Unified) & 100 & 16.3M \\ 
\bottomrule
\end{tabular}
\caption{Comparison of floating point vs integer parameters for vocabulary size of 32k, embedding size 512, 512 groups, and 50 clusters}
\label{tab:discreteparameters}
\end{table}

%\paragraph{Predicting Mean and Variance}

% \begin{table*}[t]
% \centering
% \begin{tabular}{l|cc|c|cc|ccc}
% \toprule
% \multirow{2}{*}{Model} &   \multicolumn{3}{c}{WMT En-Fr} & \multicolumn{2}{c|}{WMT En-De} & \multicolumn{3}{c}{IWSLT Pt-En} \\
% \cline{2-3} \cline{5-9}
%  & \makecell{Model \\ Params} & \makecell{BLEU} & Layers & \makecell{Model \\ Params} & \makecell{BLEU} & Layers & \makecell{Model \\ Params} & \makecell{BLEU} \\
% \midrule
% Transformer Baseline & 60.7M & 38.41 & 6 & 63M & 27.03 & 3 & 11.9M & 39.58 \\
% \midrule
% %Uni-variate & 0 &  & 0 &  & 0.5 &  \\
% RWE+linear & 44.4M & 35.76 & 6 & 44.2M & 23.04 & 3&  3.7M & 38.14 \\
% RWE+linear & 59M & 37.11 & 8  & 58.9M & 25.83 & 6 & 11.1M & \textbf{40.69} \\
% \bottomrule
% \end{tabular}
% \caption{NMT results for Uniform Distribution on a hyper-sphere (RWE), for different hyper-parameters}
% \label{tab:results_gaussian}
% \end{table*}

\begin{table*}[t]
\centering
\begin{tabular}{l|ccc|ccc|ccc}
\toprule
\multirow{2}{*}{Model} &   \multicolumn{3}{c|}{WMT En-Fr} & \multicolumn{3}{c|}{WMT En-De} & \multicolumn{3}{c}{IWSLT Pt-En} \\
\cline{2-4} \cline{5-10}
 & \makecell{Layers} & \makecell{Model \\ Params} & \makecell{BLEU} & Layers & \makecell{Model \\ Params} & \makecell{BLEU} & Layers & \makecell{Model \\ Params} & \makecell{BLEU} \\
\midrule
Transformer & 6 & 60.7M & 38.41 & 6 & 63M & 27.03 & 3 & 11.9M & 39.58 \\
\midrule
%Uni-variate & 0 &  & 0 &  & 0.5 &  \\
RWE+linear & 6 & 44.4M & 35.76 & 6 & 44.2M & 23.04 & 3&  3.7M & 38.14 \\
RWE+linear& 8 & 59M & 37.11 & 8  & 58.9M & 25.83 & 6 & 11.1M & \textbf{40.69} \\
\bottomrule
\end{tabular}
\caption{NMT results for Random Word Embeddings (RWE), for different hyper-parameters. \textbf{M} represents Millions, Transformer is the baseline model \cite{vaswani2017attention}.}
\label{tab:results_gaussian}
\end{table*}

\begin{table*}[t]
\centering
\begin{tabular}{l|cc|cc|cc}
\toprule
\multirow{2}{*}{Model} &  \multicolumn{2}{c|}{WMT En-Fr} & \multicolumn{2}{c|}{WMT En-De} & \multicolumn{2}{c}{IWSLT Pt-En} \\
\cline{2-7}
& \makecell{Emb.  Size} & \makecell{BLEU} & \makecell{Emb. Size} & \makecell{BLEU} & \makecell{Emb. Size} & \makecell{BLEU} \\
\midrule
Transformer Baseline & 62.5 MB & 39.29 & 72.3 MB & 27.38 & 31.3 MB & 39.88 \\
\midrule
PQ (Structured) & 11.82 MB & 39.95  & 13.65 MB & 27.04 & 5.9 MB & 40.33 \\
GPQ (Structured)  & 11.92 MB & 40.04 & 13.75 MB & 27.11 & 5.95 MB & 40.16 \\
PQ (Unified) & 11.7 MB & 40.02  & 13.54 MB & 26.87 & 5.86 MB & 39.34 \\
GPQ (Unified)  & 11.7 MB & \textbf{40.51} & 13.55 MB & \textbf{27.35} & 5.86 MB & \textbf{40.36} \\
% PQ (Unified) \textcolor{red}{[c=128]} & 13.67 MB & 40.47 & 15.8 MB & 27.05 & 6.8 MB & \textbf{40.39} \\
% k-Gaussian estimate (Unified) \textcolor{red}{[c=128]} & 13.67 MB & \textbf{40.65} & 15.8 MB & 26.89 & 6.8 MB & 40.36 \\
%\makecell{Multivariate (Unified) \\ + weighted} & 128 & 13.67 & 40.29 & 15.8 & 27.34 & 6.8 & 38.94 \\
\bottomrule
\end{tabular}
\caption{NMT results on Transformer baseline \cite{vaswani2017attention}, with 8 layers, groups (g) equal to the embedding size $n$, and 50 clusters (c). %\textcolor{red}{Use the last two rows of this table only for ablation study because for c=50 you can show the best results consistently. Also this setting is equivalent to Quantization and 10\% there is a chance that we get question to add a quantization baseline. We could add explanation and differenciate ourself with Quantization techniques. (we still do operations in FP. WE need to reconstruct.) Another point is it was the best to include a baseline with PQ (unified) [c=50] }
}
\label{tab:results_ngroups}
\end{table*}

\section{Experiments}

We evaluate our model for machine translation task on WMT-2014 English to French ($En\rightarrow Fr$), WMT-2014 English to German ($En\rightarrow De$), and IWSLT Portuguese to English ($Pt\rightarrow En$) datasets. We chose these pairs as they are good representatives of high, medium, and low resource language pairs.

For $En\rightarrow Fr$ experiments we used Sentence-Piece \cite{kudo2018sentencepiece} to extract a shared vocabulary of 32k sub-word units. We chose \textit{newstest2013} as our validation dataset and used \textit{newstest2014} as our test dataset. The dataset contains about 36M sentence-pairs.

For $En\rightarrow De$ experiments we use the same setup as \citet{vaswani2017attention}. The dataset contains about 4.5M sentence-pairs. We use a shared vocabulary of 37k sub-word units extracted using Sentence-Piece.

For $Pt\rightarrow En$ experiments, we replicate dataset configuration of \citet{MNMT_KD} for individual models. Specifically, the dataset contains 167k training pairs. We used a shared vocabulary of 32k subwords extracted with Sentence-Piece.

\paragraph{Evaluation} For all language pairs, we report case-sensitive BLEU score \cite{papineni2002bleu} using SacreBLEU\footnote{\url{https://github.com/mjpost/sacreBLEU}} \cite{post-2018-call}. We train for 100K steps and save a checkpoint every 5000 steps for $En\rightarrow Fr$ and $En\rightarrow De$ language pairs. For $Pt\rightarrow En$ translation we train for 50K steps and save checkpoint every 2000 steps. We select the best checkpoints based upon validation loss, and average best $5$ checkpoints for $En\rightarrow Fr$ and $En\rightarrow De$ language pairs, and average best $3$ checkpoints for $Pt\rightarrow En$. We do not use checkpoint averaging for our LSTM experiments. We use beam search with a beam width of 4 for all language pairs.

\subsection{Model and Training Details}
We use the Transformer model \cite{vaswani2017attention} as our baseline for all our experiments. We also report results on LSTM-based models to study our hypothesis for RNN based architectures. Our LSTM-based model uses multiple layers of bidirectional-LSTM, followed by a linear layer on the last layer to combine the internal states of LSTM. The decoder uses multiple layers of uni-directional LSTM, the final output of decoder uses attention on encoder output for generating words. All our models use weight sharing between the embedding matrix and the output projection layer \cite{press2016using}. 

%\textcolor{orange}{I think we can summarize the config of our used models in a table. (Orange is optional) We would save a paragraph but we need to add a table. } 
For the \textit{transformer-base} configuration, the model hidden size h is set to 512, the feed-forward hidden size $d_{\text{ff}}$ is set to 2048. We use different number of layers for our experiments, instead of the default 6 layers. For \textit{transformer-small} configuration the model hidden-size $n$ is set to 256, the feed-forward hidden size $d_{\text{ff}}$ is set to 1024. We use different number of layers for our experiments. For \textit{transformer-small}, the dropout configuration was set the same as Transformer Base. For LSTM experiments, we use 2 layers and the hidden size $n$ is set to 256 for \textit{LSTM-small}, and 7 layers and the hidden size $n$ is set to 512 for \textit{LSTM-large}. We use \textit{transformer-base} for WMT datasets, and \textit{transformer-small} for IWSLT dataset. We use \textit{LSTM-small} for IWSLT and \textit{LSTM-large} for WMT datasets. For all our experiments we set \textit{Numpy} \cite{van2011numpy} seed to $0$ and \textit{PyTorch} \cite{paszke2017automatic} seed to $3435$.

All models are optimized using Adam \cite{DBLP:journals/corr/KingmaB14} and the same learning rate schedule as proposed by \cite{vaswani2017attention}. We use label smoothing with 0.1 weight for the uniform prior distribution over the vocabulary \cite{DBLP:journals/corr/SzegedyVISW15,DBLP:journals/corr/PereyraTCKH17}. 

%\textcolor{red}{I think it would be really goof if we add a table and compare the number of embedding parameters of different techniques we are discussing in this paper. }

We train all our models on 8 NVIDIA V100 GPUs. Each training batch contains a set of sentence pairs containing approximately 6144 source tokens and 6144 target tokens for each GPU worker. For implementing our method we use the OpenNMT library \cite{opennmt}, implemented in PyTorch.

\begin{table*}[t]
\centering
\begin{tabular}{l|cc|cc|cc}
\toprule
\multirow{2}{*}{Model} &  \multicolumn{2}{c|}{WMT En-Fr} & \multicolumn{2}{c|}{WMT En-De} & \multicolumn{2}{c}{IWSLT Pt-En} \\
\cline{2-7}
& \makecell{Emb.  Size} & \makecell{BLEU} & \makecell{Emb.  Size} & \makecell{BLEU} & \makecell{Emb.  Size} & \makecell{BLEU} \\
\midrule
LSTM Baseline & 62.5 MB & 32.61 & 72.3 & 22.17 & 31.3 & 35.12 \\
\midrule
PQ (Unified) & 13.67 MB & 32.05  & 15.8 MB  & 22.13 & 6.8 MB & 35.35  \\
GPQ (Unified) & 13.67 MB & \textbf{33.82} & 15.8 MB & \textbf{22.14} & 6.8 MB & \textbf{35.58} \\
\bottomrule
\end{tabular}
\caption{NMT results on LSTM based model, we use 128 clusters (c) and groups (g) equal to embedding size $n$. 
%\textcolor{red}{Play with C and run other experiments until we get consistent result for EN-DE as well.}
}
\label{tab:results_lstm}
\end{table*}

\begin{table*}[t]
\centering
\begin{tabular}{l|ccc|ccc|ccc}
\toprule
\multirow{2}{*}{Model} &  \multicolumn{3}{c|}{WMT En-Fr} & \multicolumn{3}{c|}{WMT En-De} & \multicolumn{3}{c}{IWSLT Pt-En} \\
\cline{2-10}
& \makecell{Emb.  Size} & \makecell{BLEU} & \makecell{c} & \makecell{Emb.  Size} & BLEU & \makecell{c} & \makecell{Emb.  Size} & BLEU & \makecell{c} \\
\midrule
Transformer & 62.5 MB & 39.29 & - & 72.3 MB & 27.38 & - & 31.3 MB & 39.88 & - \\
\midrule
%\textcolor{red}{PQ (Structured) --Remove} & 3.2 MB & \textbf{40.01} & 1024 & 3.4 MB & 26.38 & 1024 & 2.2 MB & \textbf{41.45} & 1024 \\
%\textcolor{red}{GPQ--Remove} & 5.2 MB & 39.85 & 1024 & 5.4 MB & \textbf{26.48} & 1024 & 3.2 MB & 40.36 & 1024 \\
%\midrule
% Pten: 256c
PQ (Struct.) & 1.25 MB & 39.13 & 140 & 1.44 MB & 25.35 & 160 &  1.22 MB & 40.6 & 256 \\
GPQ (Struct.) & 1.25 MB & 39.18 & 102 & 1.44 MB & \textbf{26.08} & 116 & 1.1 MB & 41.1 & 128 \\
PQ (Unified) & 1.25 MB & 39.29 & 1024 & 1.44 MB & 25.83 & 1024 & 1.24 MB & 40.62 & 1024  \\
GPQ (Unified) & 1.28 MB & \textbf{39.77} & 1024 & 1.47 MB &  25.96  & 1024 & 1.25 MB & \textbf{41.32} & 1024 \\
\bottomrule
\end{tabular}
\caption{NMT results on Transformer baseline \cite{vaswani2017attention}, with 32 groups (g) and 8 layers. \textit{c} represents the number of clusters. \textit{Struct.} and \textit{unified} represents structured and unified partitioning functions respectively.}
\label{tab:results_32groups}
\end{table*}

\section{Results}

\subsection{Random Word Embeddings}
We replace the embedding matrix of the transformer model using RWE method, and summarize our results in Table \ref{tab:results_gaussian}. We evaluate our model in two cases.

In the first case, we use exactly the same hyper-parameters as our baseline model, and only replace the embedding matrix. In this case our model has only 2 embedding parameters to store the mean and variance of the Gaussian distribution, and an additional linear layer for re-scaling the embedding vectors. Our results show that without any embedding parameters, and a linear transformation, transformer model scores only $~1.44$ BLEU points below the baseline for $Pt\rightarrow En$. Deterioration in the case of $En\rightarrow Fr$ was approximately $1$ BLEU point higher than $Pt\rightarrow En$, and for $En\rightarrow De$ the drop is $~4$ BLEU points.
%of $~2.87$ BLEU points.$~2.65$

In the previous case, our model used at least 27\% fewer parameters compared to the baseline model. In order to compare RWE in a fair parameter setting, we increase the number of layers in Transformer to approximately match the total parameters with the baseline model. In this setting our model performed within $1.3$ BLEU points for $En\rightarrow De$ and $En\rightarrow Fr$. On $Pt\rightarrow En$ experiments, our model scored $~1.0$ BLEU point higher than the baseline with slightly fewer parameters. 

Overall, our experiments show that transformer based neural networks have the capacity to efficiently learn semantic and syntactic information in deeper layers, though utility of using embedding tables for token representation is beneficial and not redundant. 

\subsection{Gaussian Product Quantization}
 We apply the GPQ method on the embedding matrix of Transformer and LSTM-based models. We design a set of experiments to investigate, 1) if we can approximate the embedding matrix in an almost discrete space using integer values, 2) to study the compression ability of our method compared to PQ.

\paragraph{Experiments for Discrete Space Approximation} We evaluate our method on Transformer (Table \ref{tab:results_ngroups}) and LSTM-based models (Table \ref{tab:results_lstm}). For all experiments we chose the number of groups $g$ equal to the size of embedding vector $n$. This allowed us to maximize the number of groups and in turn maximize the amount of discrete information (integer values) and minimize the amount of floating-point parameters required to reconstruct the embedding matrix. We experiment using $50$ clusters for transformer based model and $128$ clusters for the LSTM model. Our results for the GPQ method with unified partitioning function show that with only 50 clusters for transformer model and 128 clusters for LSTM model, we are able to perform on par or better than baseline models. This implies that in all experiments we were able to approximate the embedding matrix with approximately $\approx100\%$ integer values. Specifically, if $g=n$, our \textit{unified} partitioning function used only $2c$ floating points (for storing the mean and variance clusters), while $structured$ partitioning function uses $2cn$ floating points. 
%\sout{An almost discrete embedding matrix consisting of cluster numbers, allows the sharing of floating parameters based upon the cluster assignment. Thus, our method can be implemented more efficiently, by not recomputing operations on the parameters belonging to the same cluster assignment.} 
In Table \ref{tab:results_ngroups}, 
%for all experiments the performance of our k-Gaussian estimate method is similar to PQ, with only a small variance. 
GPQ with \textit{unified} partitioning function outperforms other models consistently across all datasets. For $En\rightarrow De$, our approach is the only method that performs on par with baseline model with $5$x fewer embedding parameters. 

In Table \ref{tab:results_lstm}, we report LSTM experiments for only the \textit{unified} partitioning function as it was the best performing method. For LSTM, GPQ (\textit{Unified}) performed better than PQ (\textit{Unified}) for $En\rightarrow Fr$ with about $~1.77$ BLEU, for other languages the difference was insignificant. 
%\textcolor{red}{The drop of EN-DE is significant. I wish we had tried other settings until we fill this gap and we could report the best result.  }

\paragraph{Experiments for Compression Analysis} Our objective is to compare the capacity of GPQ method for compressing the embedding matrix, compared to PQ. We set the number of clusters to 1024 and groups to 32, as reducing the number of groups has greater affect on compression rate. We report the results in Table \ref{tab:results_32groups}. We find that with unified partition function GPQ method performs significantly better than the Transformer baseline for $En\rightarrow Fr$ (+0.48) and $Pt\rightarrow En$ (+0.7) datasets. For $En\rightarrow DE$ GPQ (unified) performs only $0.03$ BLEU points lower than the baseline. For all language pairs GPQ (unified) is the best performing model compared to PQ and other variants.

%\sout{With structured partitioning function, PQ performs significantly better (+1.09) for $Pt\rightarrow En$, this might be because unified partition function uses fewer total number of clusters, thus the higher variance in clusters is better approximated by a stochastic approach of GPQ.}

% \begin{table}[t]
% \centering
% \begin{tabular}{|l|cc|cc|}
% \toprule
% \multirow{2}{*}{Model} &  \multicolumn{2}{g|}{128 Clusters} & \multicolumn{2}{g|}{256 Groups} \\

% & \makecell{g} & \makecell{BLEU}& \makecell{c} & \makecell{BLEU} \\
% \midrule
% Baseline & - & 42.51 & - & 42.51 \\
% GM (U) & 4 & 38.46 & 4 & 39.17 \\
% GM (U) & 8 & 39.57 & 8 & 38.99 \\
% GM (U) & 16 & 39.71 & 16 & 38.53 \\
% GM (U) & 64 & 39.36 & 32 & 38.9 \\
% \bottomrule
% \end{tabular}
% \caption{NMT Results on IWSLT Pt-En, GM (U) is k-Gaussian estimate (Unified) model}
% \label{tab:ablation}
% \end{table}

\begin{figure}[t!]
\centering
  \begin{tikzpicture}
  \begin{axis}[
      title={},
      width  = 0.40*\textwidth,
      xlabel={Clusters/Groups},
      ylabel={BLEU Score},
      xmin=4, xmax=64,
      ymin=38, ymax=40,
      xtick={4,8,16,32,64},
      ytick={38,38.5,39,39.5,40},
      legend style={
                  at={(0.97,0.3)},
                  anchor=north east
          },
      ymajorgrids=true,
      grid style=dashed,
  ]

  \addlegendimage{line width=0.5pt, color=green, mark=otimes},
  \addlegendentry{128 Clusters},
  \addlegendimage{line width=0.5pt, color=purple, mark=diamond},
  \addlegendentry{256 Groups},
  
  \addplot[
      color=green,
      mark=otimes,
      ]
      coordinates {
      (4,  38.46)(8,  39.57)(16,  39.71)(64,  39.36)
      };
     
     %below is rush gigaword
    \addplot[
      color=purple,
      mark=diamond,
      ]
      coordinates {
      (4,  39.17)(8,   38.99)(16,   38.53)(32,   38.9)
      };

  \end{axis}
  \end{tikzpicture}
  \caption{\label{fig:ablation}NMT Results on IWSLT Pt-En for GPQ method with unified partitioning function}
\end{figure}
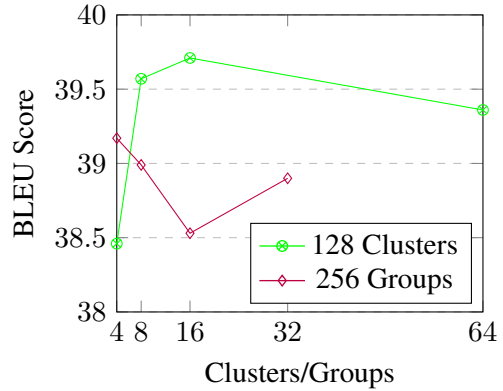

\begin{table*}[t]
\centering
\begin{tabular}{l|cc|cc}
\toprule
\multirow{2}{*}{Model} &  \multicolumn{2}{c|}{WMT En-Fr} & \multicolumn{2}{c}{WMT En-De}  \\
\cline{2-5}
& \makecell{Emb.  Size} & \makecell{BLEU} & \makecell{Emb.  Size} & \makecell{BLEU} \\
\midrule
Transformer Baseline & 1.0x & 38.41 & 1.0x & 27.03 \\
\midrule
%*Smaller Transformer Network (416) & 1.23x & 37.26 & 1.28x & 26.72  \\
*SVD with rank 64 & 7.87x & 37.44 & 7.89x & 26.32 \\
*GroupReduce \cite{chen2018groupreduce}* & 7.79x & 37.63 & 7.88x & 26.75 \\
%PQ & 7.90x & 37.78 & 7.89x & 26.34 & 3.97x & 41.27 \\
*Tensor Train \cite{khrulkov2019tensorized} & 7.72x & 37.27 & 7.75x & 26.19 \\
%Distilled Embedding & 7.87x & 37.78 & 7.89x & \textbf{26.97} & 3.96x & \textbf{42.62} \\
\midrule
GPQ (Unified) & 7.9x & \textbf{39.65} & 7.9x & \textbf{26.84} \\
\bottomrule
\end{tabular}
\caption{NMT results on transformer based models, with 256 clusters (c), 256 groups (g) and 6 layers. Results in rows marked with \textbf{*} are taken from \citet{lioutas2019distilled}}
\label{tab:compression}
\end{table*}

\section{Discussion}
\label{sec:ablation}

\subsection{Cluster and Group Size Analysis}
We study the effect of different cluster and group sizes on the GPQ method with \textit{unified} partitioning function, on the $Pt\rightarrow En$ dataset and plot results in Figure \ref{fig:ablation}. We experiment with fixing the groups to $g=256$ and varying the cluster size. We find out that with only 4 clusters and only 8 floating point parameters our model performed the best. We also experiment with fixing the clusters and varying the group size. We find that group size of 4 was the worst performing setting as a result of excessive regularization. All group sizes larger than 4 had minor influence on performance. We find that we need only 8 floating point parameters, to approximate the embedding matrix without any performance drop. Thus most important information lies in a latent space which is discrete, consisting of integer values.

\subsection{Compression Analysis}

We compare the performance of our model in a compression setting, with different compression baselines in Table \ref{tab:compression}. In particular, we compare results with two state-of-the-art methods \cite{khrulkov2019tensorized,chen2018groupreduce}. Additionally, we compare two standard baselines reported in \citet{lioutas2019distilled}. For all models we compare our results directly from \citet{lioutas2019distilled}, and choose a similar compression ratio. For both $En \rightarrow Fr$ and $En \rightarrow De$ language pairs, our GPQ (unified) model performs the best compared to all other models. Additionally, our GPQ (unified) model is the only model that performs better than the baseline model for $En\rightarrow Fr$ language pair.

%In Table \ref{tab:compression}, we compare our GPQ model with different state-of-the-art embedding compression methods. On $En\rightarrow Fr$ language pair GPQ scores $1.87$ BLEU points higher compared to the best performing compression model. Additionally, our GPQ model is the only model that performs better than the baseline model for $En\rightarrow Fr$ language pair. For $En\rightarrow De$ our method performs \textcolor{blue}{the best but} on par with other compression methods.

\subsection{Importance of Variance}
We extended PQ by introducing variance information in our GPQ method. Results in Table \ref{tab:results_lstm} highlight the importance of this information for LSTM based models. Table \ref{tab:compression} shows that incorporating variance information can be beneficial for transformer based models in a high compression setup. Lastly, Table \ref{tab:results_ngroups} shows that GPQ model is particularly beneficial when using $unified$ partitioning function. $Unified$ partitioning function uses much smaller number of total clusters, thus each cluster has a higher variance from the cluster centroid, compared to the case of $structured$ partitioning function. Thus, incorporating variance information using GPQ is beneficial, when cluster values have high variance.

% In order, to evaluate importance of variance for transformer models, for embedding matrix approximated in an almost discrete space, we report results with $n$ groups and 128 clusters in Table \ref{tab:comp_kgauss}. Our results show that variance information was beneficial for only $En \rightarrow Fr$ language-pair. \textcolor{red}{What is the difference of this table with table 3? We could how the importance of variance there as well.What is our conclusion from Table 7? If it is not a significant thing, it is better to remove that. This result put GPQ under question. }

\section{Related Work}
There are different embedding compression techniques in the literature, these methods are based on either singular value decomposition (SVD) or product quantization (PQ). In general, SVD-based methods \cite{chen2018groupreduce,khrulkov2019tensorized} approximate embedding matrices using fewer parameters by projecting high-dimensional matrices to lower dimensions. Therefore, the learning process is modified to work with new and smaller matrices where a multiplication of these matrices reconstruct original values. 
The model proposed by \citet{chen2018groupreduce} is a well-known example of SVD-based model in this field. In this technique, words are first divided into groups based upon frequency, then weighted SVD is applied on each group, which allows a compressed representation of the embedding matrix. \citet{khrulkov2019tensorized} used a multi-linear variation instead of regular SVD. The embedding dimensions are first factorized to obtain indices of smaller tensor embedding. The tensor embedding has fewer parameters and can be used to reconstruct the original embedding matrix.

% \textcolor{red}{Probabilistic embeddings: We are discussing Gaussian embedding in our paper. Is there any related work there? Are we claiming that we are better than them? \\
% https://arxiv.org/abs/1412.6623 \\
% https://arxiv.org/abs/1808.07016 \\
% and many others... \\
% }

% The below PQ intro is needed only if it is added before Background
%Product quantization based techniques\cite{jegou2010product}, first partition embedding matrix into columns and then cluster each set of columns separately. 
\citet{shu2017compressing} proposed to compress the embedding matrix by learning an integer vector for all words in vocabulary (code matrix), using the Gumbel softmax technique. The code matrix is equivalent to the quantization matrix in PQ. An embedding vector is reconstructed by looking up all the values in the code matrix from the corresponding set of vector dictionaries. The vectors are added (in PQ the vectors are concatenated) to obtain the final embedding vector.

\citet{svenstrup2017hash} proposed a hashing based technique with a common memory for storing vectors. The method is similar to \citet{shu2017compressing}, with two key differences. First, the code matrix is not learned but assigned uniformly using a hashing function. Second, instead of a simple summation of component vectors, weighted summation is applied to obtain the final embedding vector.

\cite{kim2020adaptive} learn the code matrix by utilizing binarized code learning introduced in \cite{tissier2019near}. The key novelty introduced by \citet{kim2020adaptive}, is to learn different length of code vector for each word.

\section{Conclusion}
Our work points towards the need of rethinking the process of encoding tokens to real valued vectors for machine translation. We show that Transformer model is capable of recovering useful syntactic and semantic information from a random assignment of embedding vectors. Our variant of product quantization was able to approximate the embedding matrix in an almost $~100$\% discrete space, with better performance than the baseline model. A discrete space is easier to interpret, compared to a continuous space, and can motivate future research to handle unknown tokens through optimum cluster selection.

\normalem
\bibliography{anthology,custom}
\bibliographystyle{acl_natbib}

% \appendix

% \section{Example Appendix}
% \label{sec:appendix}

% This is an appendix.

\end{document}